\documentclass[11pt,a4paper]{article}
\pdfoutput=1
\usepackage[hyperref]{acl2019}
\usepackage{times}
\usepackage{latexsym}

\usepackage{tabularx}
\usepackage{graphicx}
\usepackage{array}
\usepackage{multirow}
\usepackage{amsmath}
\usepackage{booktabs}
\usepackage{threeparttable}
\usepackage[toc,page]{appendix}
\usepackage{float}
\usepackage{makecell}
\usepackage{listings}

\usepackage{url}

\aclfinalcopy 

\title{ScispaCy: Fast and Robust Models\\ for Biomedical Natural Language Processing}

\author{\makecell{Mark Neumann, Daniel King, Iz Beltagy, Waleed Ammar} \\
Allen Institute for Artificial Intelligence, Seattle, WA, USA\\
{\tt $\{$markn,daniel,beltagy,waleeda$\}$@allenai.org}\\}

\date{}

\newcommand\wacomment[1]{\textcolor{blue}{[#1] --\textsc{WA}}}

\begin{document}
\maketitle
\begin{abstract}
Despite recent advances in natural language processing, many statistical models for processing text perform extremely poorly under domain shift. Processing biomedical and clinical text is a critically important application area of natural language processing, for which there are few robust, practical, publicly available models. This paper describes scispaCy, a new Python library and models for practical biomedical/scientific text processing, which heavily leverages the spaCy library. We detail the performance of two packages of models released in scispaCy and demonstrate their robustness on several tasks and datasets. Models and code are available at \url{https://allenai.github.io/scispacy/}.
\end{abstract}

\section{Introduction}

The publication rate in the medical and biomedical sciences is growing at an exponential rate \citep{DBLP:journals/corr/BornmannM14}. The information overload problem is widespread across academia, but is particularly apparent in the biomedical sciences, where individual papers may contain specific discoveries relating to a dizzying variety of genes, drugs, and proteins. In order to cope with the sheer volume of new scientific knowledge, there have been many attempts to automate the process of extracting entities, relations, protein interactions and other structured knowledge from scientific papers \citep{Wei2016AssessingTS, Ammar2018ConstructionOT, Poon2014LiteromePG}.

\begin{figure}
    \centering
    \includegraphics[width=0.5\textwidth]{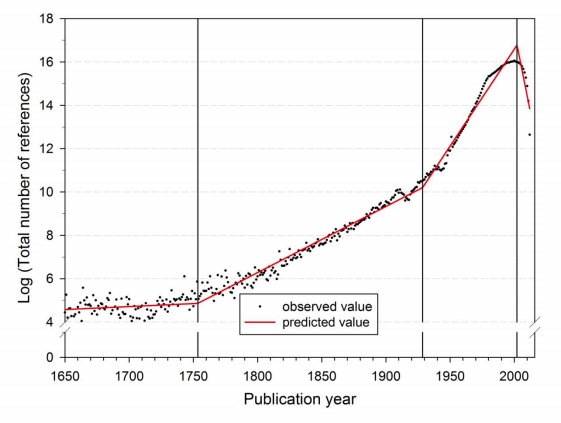}
    \caption{Growth of the annual number of cited references from 1650 to 2012 in
the medical and health sciences (citing publications from 1980 to 2012). Figure from \cite{DBLP:journals/corr/BornmannM14}.}
    \label{fig:my_label}
\end{figure}

Although there exists a wealth of tools for processing biomedical text, many focus primarily on named entity recognition and disambiguation. MetaMap and MetaMapLite \citep{Aronson2001EffectiveMO, DemnerFushman2017MetaMapLA}, the two most widely used and supported tools for biomedical text processing, support entity linking with negation detection and acronym resolution. However, tools which cover more classical natural language processing (NLP) tasks such as the GENIA tagger \citep{Tsuruoka2005DevelopingAR, Tsuruoka2005BidirectionalIW}, or phrase structure parsers such as those presented in \citet{McClosky2008SelfTrainingFB} typically do not make use of new research innovations such as word representations or neural networks.

In this paper, we introduce scispaCy, a specialized NLP library for processing biomedical texts which builds on the robust spaCy library,\footnote{\url{spacy.io}} and document its performance relative to state of the art models for part of speech (POS) tagging, dependency parsing, named entity recognition (NER) and sentence segmentation. Specifically, we:

\begin{itemize}
    \item Release a reformatted version of the GENIA 1.0 \citep{Kim2003GENIAC} corpus converted into Universal Dependencies v1.0 and aligned with the original text from the PubMed abstracts.
    \item Benchmark 9 named entity recognition models for more specific entity extraction applications demonstrating competitive performance when compared to strong baselines.
    \item Release and evaluate two fast and convenient pipelines for biomedical text, which include tokenization, part of speech tagging, dependency parsing and named entity recognition.
\end{itemize}

\section{Overview of (sci)spaCy}\label{sec:overview}
In this section, we briefly describe the models used in the spaCy library and describe how we build on them in scispaCy. 

\paragraph{spaCy.}
The Python-based spaCy library \citep{spacy2}\footnote{Source code at \url{https://github.com/explosion/spaCy}} provides a variety of practical tools for text processing in multiple languages. Their models have emerged as the defacto standard for practical NLP due to their speed, robustness and close to state of the art performance. As the spaCy models are popular and the spaCy API is widely known to many potential users, we choose to build upon the spaCy library for creating a biomedical text processing pipeline.

\paragraph{scispaCy.}
Our goal is to develop scispaCy as a robust, efficient and performant NLP library to satisfy the primary text processing needs in the biomedical domain.
In this release of scispaCy, we retrain spaCy\footnote{scispaCy models are based on spaCy version 2.0.18} models for POS tagging, dependency parsing, and NER using datasets relevant to biomedical text, and enhance the tokenization module with additional rules.
scispaCy contains two core released packages: \textbf{\lstinline{en_core_sci_sm}} and \textbf{\lstinline{en_core_sci_md}}. Models in the \textbf{\lstinline{en_core_sci_md}} package have a larger vocabulary and include word vectors, while those in \textbf{\lstinline{en_core_sci_sm}} have a smaller vocabulary and do not include word vectors, as shown in Table \ref{vocab}.

\begin{table}
\centering
\small
\setlength\tabcolsep{3pt}
\begin{tabular}{@{}lp{1.5cm}p{1.0cm}p{1.0cm}p{1.0cm}@{}}
\toprule
Model        & Vocab Size & Vector Count & Min Word Freq & Min Doc Freq \\ \midrule
\textbf{\lstinline{en_core_sci_sm}} & 58,338          & 0            & 50            & 5            \\
\textbf{\lstinline{en_core_sci_md}} & 101,678         & 98,131       & 20            & 5            \\ \bottomrule
\end{tabular}
\caption{Vocabulary statistics for the two core packages in scispaCy.}
\label{vocab}
\end{table}

\paragraph{Processing Speed.} 
To emphasize the efficiency and practical utility of the end-to-end pipeline provided by scispaCy packages, we perform a speed comparison with several other publicly available processing pipelines for biomedical text using 10k randomly selected PubMed abstracts.
We report results with and without segmenting the abstracts into sentences since some of the libraries (e.g., GENIA tagger) are designed to operate on sentences. 

\begin{table}[t]
\small
\setlength\tabcolsep{3pt}
\begin{tabular}{r|rr}
\toprule
& \multicolumn{2}{c}{Processing Times Per} \\ 
Software Package & Abstract (ms) & Sentence (ms) \\ \midrule
NLP4J (java)                        & 19    & 2      \\
Genia Tagger (c++)                  & 73    & 3      \\
Biaffine (TF)                       & 272     & 29    \\
Biaffine (TF + 12 CPUs)             & 72 &   7 \\
jPTDP (Dynet)                       & 905 & 97  \\
Dexter v2.1.0 &                      208 & 84 \\
MetaMapLite v3.6.2                  & 293 & 89 \\ \midrule
\textbf{\lstinline{en_core_sci_sm}} & 32    & 4     \\
\textbf{\lstinline{en_core_sci_md}} & 33 & 4    \\ \bottomrule
\end{tabular}
\caption{Wall clock comparison of different publicly available biomedical NLP pipelines. All experiments run on a single machine with 12 Intel(R) Core(TM) i7-6850K CPU @ 3.60GHz and 62GB RAM. For the Biaffine Parser, a pre-compiled Tensorflow binary with support for AVX2 instructions was used in a good faith attempt to optimize the implementation. Dynet does support the Intel MKL, but requires compilation from scratch and as such, does not represent an ``off the shelf" system. TF is short for Tensorflow.}
\label{speed}
\end{table}

As shown in Table \ref{speed}, both models released in scispaCy demonstrate competitive speed to pipelines written in C++ and Java, languages designed for production settings.

Whilst scispaCy is not as fast as pipelines designed for purely production use-cases (e.g., NLP4J), it has the benefit of straightforward integration with the large ecosystem of Python libraries for machine learning and text processing. Although the comparison in Table \ref{speed} is not an apples to apples comparison with other frameworks (different tasks, implementation languages etc), it is useful to understand scispaCy's runtime in the context of other pipeline components. Running scispaCy models \textit{in addition to} standard Entity Linking software such as MetaMap would result in only a marginal increase in overall runtime.

In the following section, we describe the POS taggers and dependency parsers in scispaCy.

\section{POS Tagging and Dependency Parsing}\label{sec:syntax}
The joint POS tagging and dependency parsing model in spaCy is an arc-eager transition-based parser trained with a dynamic oracle, similar to \citet{Goldberg2012ADO}. 
Features are CNN representations of token features and shared across all pipeline models \cite{Kiperwasser2016SimpleAA, Zhang2016StackpropagationIR}.
Next, we describe the data we used to train it in scispaCy.

\subsection{Datasets}\label{sec:genia}
\paragraph{GENIA 1.0 Dependencies.}

To train the dependency parser and part of speech tagger in both released models, we convert the treebank of \citet{McClosky2008SelfTrainingFB},\footnote{\url{https://nlp.stanford.edu/~mcclosky/biomedical.html}} which is based on the GENIA 1.0 corpus \citep{Kim2003GENIAC}, to Universal Dependencies v1.0 using the Stanford Dependency Converter \citep{Schuster2016EnhancedEU}. 
As this dataset has POS tags annotated, we use it to train the POS tagger jointly with the dependency parser in both released models. 

As we believe the Universal Dependencies converted from the original GENIA 1.0 corpus are generally useful, we have released them as a separate contribution of this paper.\footnote{\url{https://github.com/allenai/genia-dependency-trees}}
In this data release, we also align the converted dependency parses to their original text spans in the raw, untokenized abstracts from the original release,\footnote{Available at \url{http://www.geniaproject.org/}} 
and include the PubMed metadata for the abstracts which was discarded in the GENIA corpus released by \citet{McClosky2008SelfTrainingFB}.
We hope that this raw format can emerge as a resource for practical evaluation in the biomedical domain of core NLP tasks such as tokenization, sentence segmentation and joint models of syntax. 

Finally, we also retrieve from PubMed the original metadata associated with each abstract. This includes relevant named entities linked to their Medical Subject Headings (MeSH terms) as well as chemicals and drugs linked to a variety of ontologies, as well as author metadata, publication dates, citation statistics and journal metadata. We hope that the community can find interesting problems for which such natural supervision can be used.

\begin{table}[t]
\centering
\begin{tabular}{@{}lc@{}}
\toprule
Package/Model                  & GENIA \\ \midrule
MarMoT                 & 98.61      \\
jPTDP-v1               & 98.66      \\
NLP4J-POS              & 98.80       \\
BiLSTM-CRF             & 98.44      \\
BiLSTM-CRF- charcnn    & 98.89      \\
BiLSTM-CRF - char lstm & 98.85      \\
 \midrule
\textbf{\lstinline{en_core_sci_sm}} & 98.38       \\
\textbf{\lstinline{en_core_sci_md}} & 98.51      \\
\bottomrule
\end{tabular}
\caption{Part of Speech tagging results on the GENIA Test set.}
\label{pos-tagging}
\end{table}

\paragraph{OntoNotes 5.0.}
To increase the robustness of the dependency parser and POS tagger to generic text, we make use of the OntoNotes 5.0 corpus\footnote{Instructions for download at \url{http://cemantix.org/data/ontonotes.html}} when training the dependency parser and part of speech tagger \citep{Weischedel2011OntoNotes, Hovy2006OntoNotesT9}. The OntoNotes corpus consists of multiple genres of text, annotated with syntactic and semantic information, but we only use POS and dependency parsing annotations in this work.

\begin{table}[t]
\begin{tabular}{@{}lcc@{}}
\toprule
Package/Model                          & UAS   & LAS   \\  \midrule
Stanford-NNdep                 & 89.02 & 87.56 \\
NLP4J-dep                      & 90.25 & 88.87 \\
jPTDP-v1                       & 91.89 & 90.27 \\
Stanford-Biaffine-v2           & 92.64 & 91.23 \\
Stanford-Biaffine-v2(Gold POS) & 92.84 & 91.92 \\ \midrule
\textbf{\lstinline{en_core_sci_sm} - SD} & 90.31 & 88.65  \\
\textbf{\lstinline{en_core_sci_md} - SD} & 90.66 & 88.98 \\ \midrule
\textbf{\lstinline{en_core_sci_sm}} & 89.69 & 87.67  \\
\textbf{\lstinline{en_core_sci_md}} & 90.60 & 88.79 \\ \bottomrule
\end{tabular}
\caption{Dependency Parsing results on the GENIA 1.0 corpus converted to dependencies using the Stanford Universal Dependency Converter. We additionally provide evaluations using Stanford Dependencies(SD) in order for comparison relative to the results reported in \citep{NguyenVerspoor2018bionlp}.}
\label{dependencies}
\end{table}

\subsection{Experiments}\label{sec:syntax_experiments}

We compare our models to the recent survey study of dependency parsing and POS tagging for biomedical data \citep{NguyenVerspoor2018bionlp} in Tables \ref{pos-tagging} and \ref{dependencies}. POS tagging results show that both models released in scispaCy are competitive with state of the art systems, and can be considered of equivalent practical value. In the case of dependency parsing, we find that the Biaffine parser of \citet{Dozat2016DeepBA} outperforms the scispaCy models by a margin of 2-3\%. However, as demonstrated in Table \ref{speed}, the scispaCy models are approximately 9x faster due to the speed optimizations in spaCy. \footnote{We refer the interested reader to \citet{NguyenVerspoor2018bionlp} for a comprehensive description of model architectures considered in this evaluation.}

\paragraph{Robustness to Web Data.}

\begin{figure}[ht]
    \includegraphics[width=0.45\textwidth]{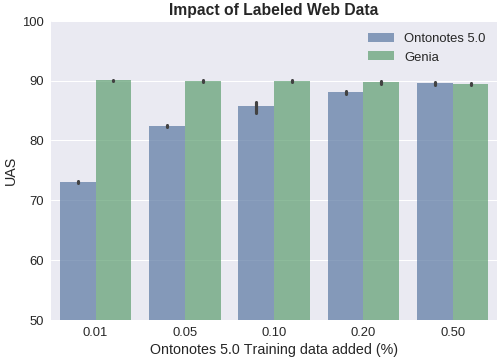}
    \caption{Unlabeled attachment score (UAS) performance for an \textbf{\lstinline{en_core_sci_md}} model trained with increasing amounts of web data incorporated. Table shows mean of 3 random seeds.}
    \label{fig:ontonotes_mix}
\end{figure}

A core principle of the scispaCy models is that they are useful on a wide variety of types of text with a biomedical focus, such as clinical notes, academic papers, clinical trials reports and medical records. In order to make our models robust across a wider range of domains more generally, we experiment with incorporating training data from the OntoNotes 5.0 corpus when training the dependency parser and POS tagger. Figure \ref{fig:ontonotes_mix} demonstrates the effectiveness of adding increasing percentages of web data, showing substantially improved performance on OntoNotes, at no reduction in performance on biomedical text. Note that mixing in web text during training has been applied to previous systems - the GENIA Tagger \citep{Tsuruoka2005DevelopingAR} also employs this technique.

\section{Named Entity Recognition}
 The NER model in spaCy is a transition-based system based on the chunking model from \citet{Lample2016NeuralAF}. Tokens are represented as hashed, embedded representations of the prefix, suffix, shape and lemmatized features of individual words. 
 Next, we describe the data we used to train NER models in scispaCy.

\subsection{Datasets}

The main NER model in both released packages in scispaCy is trained on the mention spans in the MedMentions dataset \citep{Murty2018HierarchicalLA}. Since the MedMentions dataset was originally designed for entity linking, this model recognizes a wide variety of entity types, as well as non-standard syntactic phrases such as verbs and modifiers, but the model does not predict the entity type.
In order to provide for users with more specific requirements around entity types, we release four additional packages \textbf{\lstinline{en_ner_\{bc5cdr|craft|jnlpba|bionlp13cg\}_md}} with finer-grained NER models trained on BC5CDR \cite[for chemicals and diseases;][]{Li2016BioCreativeVC}, CRAFT \citep[for cell types, chemicals, proteins, genes;][]{Bada2011ConceptAI}, JNLPBA \cite[for cell lines, cell types, DNAs, RNAs, proteins;][]{Collier2004IntroductionTT} and BioNLP13CG \cite[for cancer genetics;][]{Pyysalo2015OverviewOT}, respectively.

\subsection{Experiments}
\label{sec:ner_experiments}
As NER is a key task for other biomedical text processing tasks, we conduct a through evaluation of the suitability of scispaCy to provide baseline performance across a wide variety of datasets. In particular, we retrain the spaCy NER model on each of the four datasets mentioned earlier (BC5CDR, CRAFT, JNLPBA, BioNLP13CG) as well as five more datasets in \citet{Crichton2017ANN}: AnatEM, BC2GM, BC4CHEMD, Linnaeus, NCBI-Disease. 
These datasets cover a wide variety of entity types required by different biomedical domains, including cancer genetics, disease-drug interactions, pathway analysis and trial population extraction. Additionally, they vary considerably in size and number of entities. For example, BC4CHEMD \citep{Krallinger2015CHEMDNERTD} has 84,310 annotations while Linnaeus \citep{Gerner2009LINNAEUSAS} only has 4,263. BioNLP13CG \citep{Pyysalo2015OverviewOT} annotates 16 entity types while five of the datasets only annotate a single entity type.\footnote{For a detailed discussion of the datasets and their creation, we refer the reader to \url{https://github.com/cambridgeltl/MTL-Bioinformatics-2016/blob/master/Additional\%20file\%201.pdf}}

Table \ref{ner} provides a thorough comparison of the scispaCy NER models compared to a variety of models. In particular, we compare the models to strong baselines which do not consider the use of 1) multi-task learning across multiple datasets and 2) semi-supervised learning via large pretrained language models. Overall, we find that the scispaCy models are competitive baselines for 5 of the 9 datasets.

Additionally, in Table \ref{tab:ner_recall} we evaluate the recall of the pipeline mention detector available in both scispaCy models (trained on the MedMentions dataset) against all 9 specialised NER datasets. Overall, we observe a modest drop in average recall when compared directly to the MedMentions results in Table \ref{tab:med_mentions}, but considering the diverse domains of the 9 specialised NER datasets, achieving this level of recall across datasets is already non-trivial.


\begin{table*}[t]
\centering
\begin{threeparttable}

\begin{tabular}{@{}l|lll|ll@{}}
\toprule
Dataset     & Baseline & SOTA & + Resources & \textbf{\lstinline{sci_sm}}    & \textbf{\lstinline{sci_md}}    \\ \midrule
BC5CDR \citep{Li2016BioCreativeVC}           & 83.87    & 86.92b       & 89.69\textsuperscript{bb}  & 78.83  & 83.92            \\
CRAFT \citep{Bada2011ConceptAI}          & 79.55     & -       &  -   & 72.31  & 76.17                         \\
JNLPBA \citep{Collier2004IntroductionTT}         & 68.95     & 73.48\textsuperscript{b}     & 75.50\textsuperscript{bb}  & 71.78  & 73.21          \\
BioNLP13CG \citep{Pyysalo2015OverviewOT} & 76.74    & -        &  -   & 72.98  & 77.60                        \\
AnatEM \citep{Pyysalo2014AnatomicalEM}   & 88.55   & 91.61\textsuperscript{**}       & - & 80.13  & 84.14                       \\
BC2GM \citep{Smith2008OverviewOB}              & 84.41   & 80.51\textsuperscript{b}         & 81.69\textsuperscript{bb}      & 75.77  & 78.30      \\
BC4CHEMD \citep{Krallinger2015CHEMDNERTD}       & 82.32     & 88.75\textsuperscript{a}      & 89.37\textsuperscript{aa}     & 82.24  & 84.55       \\
Linnaeus \citep{Gerner2009LINNAEUSAS}         & 79.33   & 95.68\textsuperscript{**}         & -      & 79.20  & 81.74           \\
NCBI-Disease \citep{Dogan2014NCBIDC}       & 77.82   & 85.80\textsuperscript{b}          & 87.34\textsuperscript{bb}      & 79.50  & 81.65         \\ \bottomrule
\end{tabular}
\begin{tablenotes}\footnotesize
\item
\textbf{bb}: LM model from \citet{Sachan2017EffectiveUO}
\textbf{b}: LSTM model from \citet{Sachan2017EffectiveUO} \\
\textbf{a}: Single Task model from \citet{Wang2018CrosstypeBN}
\textbf{aa}: Multi-task model from \citet{Wang2018CrosstypeBN} \\
\textbf{**} Evaluations use dictionaries developed without a clear train/test split.
\end{tablenotes}

\caption{Test F1 Measure on NER for the small and medium scispaCy models compared to a variety of strong baselines and state of the art models. The \textbf{Baseline} and \textbf{SOTA} (State of the Art) columns include only single models which do not use additional resources, such as language models, or additional sources of supervision, such as multi-task learning. \textbf{+ Resources} allows any type of supervision or pretraining. All scispaCy results are the mean of 5 random seeds.}
\label{ner}
\end{threeparttable}
\end{table*}

\begin{table}[H]
\centering
\begin{tabular}{@{}lcc@{}}
\toprule
Dataset & \textbf{\lstinline{sci_sm}} & \textbf{\lstinline{sci_md}}  \\ \midrule
BC5CDR                      &   75.62     &  78.79  \\
CRAFT                       &   58.28     &  58.03     \\
JNLPBA                      &   67.33     &  70.36     \\
BioNLP13CG                  &   58.93     &  60.25    \\
AnatEM                      &   56.55     &  57.94  \\
BC2GM                       &   54.87     &  56.89         \\
BC4CHEMD                    &   60.60     &  60.75   \\
Linnaeus                    &   67.48     &  68.61    \\
NCBI-Disease                &   65.76     &  65.65    \\
\textbf{Average}               &   62.81     &  64.14 \\  \bottomrule
\end{tabular}
\caption{Recall on the test sets of 9 specialist NER datasets, when the base mention detector  is trained on MedMentions. The base mention detector is available in both \textbf{\lstinline{en_core_sci_sm}} and \textbf{\lstinline{en_core_sci_md}} models. \label{tab:ner_recall}}
\end{table}

\begin{table}[H]
\centering
\begin{tabular}{@{}lccc@{}}
\toprule
Model & Precision & Recall  & F1 \\ \midrule
\textbf{\lstinline{en_core_sci_sm}} & 69.22   & 67.19 & 68.19   \\
\textbf{\lstinline{en_core_sci_md}} & 70.44   & 67.56 & 68.97 \\

\end{tabular}
\caption{Performance of the base mention detector on the MedMentions Corpus.}
\label{tab:med_mentions}
\end{table}

\section{Candidate Generation for Entity Linking}

In addition to Named Entity Recognition, scispaCy contains some initial groundwork needed to build an Entity Linking model designed to link to a subset of the Unified Medical Language System \citep[UMLS;][]{Bodenreider2004TheUM}. This reduced subset is comprised of sections 0, 1, 2 and 9 (SNOMED) of the UMLS 2017 AA release, which are publicly distributable. It contains 2.78M unique concepts and covers 99\% of the mention concepts present in the MedMentions dataset \citep{Murty2018HierarchicalLA}.

\subsection{Candidate Generation}

To generate candidate entities for linking a given mention, we use an approximate nearest neighbours search over our subset of UMLS concepts and concept aliases and output the entities associated with the nearest K. Concepts and aliases are encoded using the vector of TF-IDF scores of character 3-grams which appears in 10 or more entity names or aliases (i.e., document frequency $\ge$ 10). In total, all data associated with the candidate generator including cached  vectors for 2.78M concepts occupies 1.1GB of space on disk.

\paragraph{Aliases.}
Canonical concepts in UMLS have \textit{aliases} - common names of drugs, alternative spellings, and otherwise words or phrases that are often linked to a given concept. Importantly, aliases may be shared across concepts, such as ``cancer" for the canonical concepts of both ``Lung Cancer" and ``Breast Cancer".
Since the nearest neighbor search is based on the surface forms, it returns K string values.
However, because a given string may be an alias for multiple concepts, the list of K nearest neighbor strings may not translate to a list of K candidate entities.
This is the correct implementation in practice, because given a possibly ambiguous alias, it is beneficial to score all plausible concepts, but it does mean that we cannot determine the exact number of candidate entities that will be generated for a given value of K. 
In practice, the number of retrieved candidates for a given K is much lower than K itself, with the exception of a few long tail aliases, which are aliases for a large number of concepts. For example, for K=100, we retrieve $54.26\pm 12.45$ candidates, with the max number of candidates for a single mention being 164.

\begin{figure}
    \includegraphics[width=0.50\textwidth]{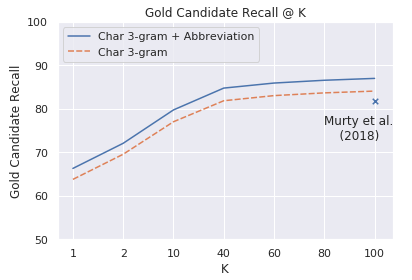}
    \caption{Gold Candidate Generation Recall for different values of K. Note that K refers to the number of nearest neighbour queries, and not the number of considered candidates. \citet{Murty2018HierarchicalLA} do not report this distinction, but for a given K the same amount of work is done (retrieving K neighbours from the index), so results are comparable. For all K, the actual number of candidates is considerably lower on average.}
    \label{fig:candidates}
\end{figure}

\paragraph{Abbreviations.}
During development of the candidate generator, we noticed that abbreviated mentions account for a substantial proportion of the failure cases where none of the generated candidates match the correct entity. To partially remedy this, we implement the unsupervised abbreviation detection algorithm of \citet{Schwartz2002ASA}, substituting mention candidates marked as abbreviations for their long form definitions before searching for their nearest neighbours. Figure \ref{fig:candidates} demonstrates the improved recall of gold concepts for various values of K nearest neighbours. Our candidate generator provides a 5\% absolute improvement over \citet{Murty2018HierarchicalLA} despite generating 46\% fewer candidates per mention on average.

\section{Sentence Segmentation and Citation Handling}

Accurate sentence segmentation is required for many practical applications of natural language processing. Biomedical data presents many difficulties for standard sentence segmentation algorithms: abbreviated names and noun compounds containing punctuation are more common, whilst the wide range of citation styles can easily be misidentified as sentence boundaries.

We evaluate sentence segmentation using both sentence and full-abstract accuracy when segmenting PubMed abstracts from the raw, untokenized GENIA development set (the \textbf{Sent/Abstract} columns in Table \ref{sent-segment}).

Additionally, we examine the ability of the segmentation learned by our model to generalise to the body text of PubMed articles. Body text is typically more complex than abstract text, but in particular, it contains citations, which are considerably less frequent in abstract text. In order to examine the effectiveness of our models in this scenario, we design the following synthetic experiment. Given sentences from \citet{Cohan2019StructuralSF} which were originally designed for citation intent prediction, we run these sentences individually through our models. As we know that these sentences should be single sentences, we can simply count the frequency with which our models segment the individual sentences containing citations into multiple sentences (the \textbf{Citation} column in Table \ref{sent-segment}).

As demonstrated by Table \ref{sent-segment}, training the dependency parser on in-domain data (both the scispaCy models) completely obviates the need for rule-based sentence segmentation. This is a positive result - rule based sentence segmentation is a brittle, time consuming process, which we have replaced with a domain specific version of an existing pipeline component.

Both scispaCy models are released with the custom tokeniser, but without a custom sentence segmenter by default. 

\begin{table}[H]
\centering
\setlength\tabcolsep{3pt}
\begin{tabular}{@{}lccc@{}}
\toprule
Model                       & Sent    & Abstract & Citation \\ \midrule
web-small          & 88.2\% & 67.5\%  & 74.4\%  \\
web-small + ct      & 86.6\% & 62.1\%  & 88.6\%  \\
web-small + cs      & 91.9\% & 77.0\%  & 87.5\%  \\
web-small + cs + ct & 92.1\% & 78.3\%  & 94.7\%  \\ \midrule
sci-small + ct      & 97.2\% & 81.7\%  & 97.9\%  \\
sci-small + cs + ct & 97.2\% & 81.7\%  & 98.0\%  \\
sci-med + ct      & 97.3\% & 81.7\%  & 98.0\%  \\
sci-med + cs + ct & 97.4\% & 81.7\%  & 98.0\%  \\ \bottomrule
\end{tabular}
\caption{Sentence segmentation performance for the core spaCy and scispaCy models. \textbf{cs} $=$ custom rule based sentence segmenter and \textbf{ct} $=$ custom rule based tokenizer, both designed explicitly to handle citations and common patterns in biomedical text.}
\label{sent-segment}
\end{table}

\section{Related Work}

Apache cTakes \citep{Savova2010MayoCT} was designed specifically for clinical notes rather than the broader biomedical domain. MetaMap and MetaMapLite \citep{Aronson2001EffectiveMO, DemnerFushman2017MetaMapLA} from the National Library of Medicine focus specifically on entity linking using the Unified Medical Language System (UMLS) \citep{Bodenreider2004TheUM} as a knowledge base. \citet{Buyko_automaticallyadapting} adapt Apache OpenNLP using the GENIA corpus, but their system is not openly available and is less suitable for modern, Python-based workflows. The GENIA Tagger \cite{Tsuruoka2005DevelopingAR} provides the closest comparison to scispaCy due to it's multi-stage pipeline, integrated research contributions and production quality runtime. We improve on the GENIA Tagger by adding a full dependency parser rather than just noun chunking, as well as improved results for NER without compromising significantly on speed.

In more fundamental NLP research, the GENIA corpus \citep{Kim2003GENIAC} has been widely used to evaluate transfer learning and domain adaptation. \citet{McClosky2006RerankingAS} demonstrate the effectiveness of self-training and parse re-ranking for domain adaptation. \citet{Rimell2008AdaptingAL} adapt a CCG parser using only POS and lexical categories, while \citet{Joshi2018ExtendingAP} extend a neural phrase structure parser trained on web text to the biomedical domain with a small number of partially annotated examples. These papers focus mainly of the problem of domain adaptation itself, rather than the objective of obtaining a robust, high-performance parser using existing resources.

NLP techniques, and in particular, \textit{distant supervision} have been employed to assist the curation of large, structured biomedical resources. \citet{Poon2015DistantSF} extract 1.5 million cancer pathway interactions from PubMed abstracts, leading to the development of Literome \citep{Poon2014LiteromePG}, a search engine for genic pathway interactions and genotype-phenotype interactions. A fundamental aspect of \citet{ValenzuelaEscarcega2018LargescaleAM} and  \citet{Poon2014LiteromePG} is the use of hand-written rules and triggers for events based on dependency tree paths; the connection to the application of scispaCy is quite apparent.

\section{Conclusion}
In this paper we presented several robust model pipelines for a variety of natural language processing tasks focused on biomedical text. The scispaCy models are fast, easy to use, scalable, and achieve close to state of the art performance. We hope that the release of these models enables new applications in biomedical information extraction whilst making it easy to leverage high quality syntactic annotation for downstream tasks. Additionally, we released a reformatted GENIA 1.0 corpus augmented with automatically produced Universal Dependency annotations and recovered and aligned original abstract metadata. Future work on scispaCy will include a more fully featured entity linker built from the current candidate generation work, as well as other pipeline components such as negation detection commonly used in the clinical and biomedical natural language processing communities.

\bibliography{acl2019}
\bibliographystyle{acl_natbib}

\end{document}